%% file: acl_latex.tex
\pdfoutput=1

\documentclass[11pt]{article}

\usepackage{acl}

\usepackage{times}
\usepackage{latexsym}
\usepackage{threeparttable}
\usepackage{tabularx} 
\usepackage[T1]{fontenc}
\usepackage{booktabs}
\usepackage{multirow}
\usepackage{amsmath}
\usepackage{geometry}
\usepackage{tcolorbox}
\usepackage{adjustbox}
\usepackage{caption}
\usepackage{graphicx}
\usepackage{xcolor}
\usepackage{tikz}

\usepackage{longtable}
\usepackage{array}
\usepackage{tabularx}
\usepackage{booktabs}

\usepackage[utf8]{inputenc}

\usepackage{microtype}

\usepackage{inconsolata}

\usepackage{graphicx}

\title{Exploring LLMs for Predicting Tutor Strategy and Student Outcomes in Dialogues}

\author{Fareya Ikram, Alexander Scarlatos, Andrew Lan\\University of Massachusetts Amherst \\ \{fikram, ajscarlatos, andrewlan\}@umass.edu} 

\begin{document}
\maketitle 

\begin{abstract}
Tutoring dialogues have gained significant attention in recent years, given the prominence of online learning and the emerging tutoring abilities of artificial intelligence (AI) agents powered by large language models (LLMs). Recent studies have shown that the strategies used by tutors can have significant effects on student outcomes, necessitating methods to predict how tutors will behave and how their actions impact students. However, few works have studied predicting tutor strategy in dialogues. Therefore, in this work we investigate the ability of modern LLMs, particularly Llama 3 and GPT-4o, to predict both future tutor moves and student outcomes in dialogues, using two math tutoring dialogue datasets. We find that even state-of-the-art LLMs struggle to predict future tutor strategy while tutor strategy is highly indicative of student outcomes, outlining a need for more powerful methods to approach this task.
\end{abstract}

\section{Introduction}

Tutoring has been shown to be highly effective in increasing student learning, both when administered by human tutors or intelligent tutoring systems \cite{metaat,nye2014autotutor}. Recently, several automated tutors, powered by large language models (LLMs), have been deployed in educational settings, such as Khan Academy's Khanmigo \cite{khanmigo} or Carnegie Learning's LiveHint \cite{livehint}.
To ensure that students benefit from these tools, it is important to study the strategies used by tutors and how they impact student learning outcomes.

Tutor strategy is commonly formalized using ``moves'', or high-level pedagogical actions taken in any given dialogue turn to support student learning~\cite{ncte,macina-etal-2023-mathdial,suresh-etal-2022-talkmoves}. Recent studies have shown that explicit move and strategy information can be used to improve tutor effectiveness \cite{wang2024bridging,wang2024tutor}. Others that train LLMs to be effective tutors \cite{tack-etal-2023-bea,learnlmteam2024learnlmimprovinggeminilearning,sonkar-etal-2024-pedagogical,huber-etal-2023-enhancing,vasselli-etal-2023-naisteacher,scarlatos2025trainingllmbasedtutorsimprove} have also highlighted the importance of pedagogical strategy. Several prior works have used tutor moves to predict student outcomes \cite{lin2022good,2024.EDM-long-papers.10,abdelshiheed2024aligning,10.1145/3706468.3706474}, though text alone processed by LLMs is often sufficient \cite{scarlatos2024exploring,10.1145/3627673.3679108}. In this work, we explicitly study the effect of move annotations, compared to text alone, on predicting tutor strategy and student outcomes.

While many works have studied how to \textit{identify} tutor moves in dialogues \cite{demszky2021measuring,wang-demszky-2024-edu,10.1145/3657604.3664664,10.1007/978-3-031-36272-9_53,mcnichols2025studychatdatasetstudentdialogues}, there are few works studying how to predict \textit{future} tutor moves. One prior work does so using GRUs and RoBERTa \cite{ganesh-etal-2021-teacher}, though to the best of our knowledge, none have studied how generative language models, such as Meta's Llama 3 \cite{dubey2024llama} or OpenAI's GPT-4 \cite{openai2024gpt4o}, perform on this task.

In this work, to address the needs of understanding tutor strategy and its effect on student outcomes, we seek to answer the following research questions: \textbf{RQ1}: Can LLMs predict \textit{tutor strategy} using tutor moves and dialogue history?, \textbf{RQ2}: Can LLMs predict \textit{student outcomes} using tutor moves and dialogue history?, and \textbf{RQ3}: Which tutoring strategies have the highest impact on student outcomes?
To the best of our knowledge, our study is the first to jointly address these questions using modern generative LLMs.
Overall, we find that tutor strategy prediction is highly challenging, with student outcome prediction being easier and facilitated by tutor move annotations. Our findings indicate that tutor strategy prediction is an important and challenging task worth further study in future work.


\section{Methodology}

\input{tables/dialogue_example}

In this section, we detail the four primary tasks we use to study tutor strategy and student outcome prediction, as well as the various LLM and non-LLM methods we evaluate on these tasks. First, we define our notation. Each tutor-student dialogue, $D=\{T_i,S_i, M_i, E_i, C\}_{i=1}^{N}$, contains a sequence of alternating textual student utterances $S_i$ and tutor utterances $T_i$. Each tutor turn is associated with one or more categorical moves, i.e., granular pedagogical actions, $M_i$, and each student turn is optionally associated with a binary measure of success $E_i$. Each dialogue is also labeled with a final binary measure of success, $C$.
As our focus is to study tutor strategy and student outcomes, rather than student behavior, we do not use student move labels.
We show an example dialogue in Table \ref{tab:dialogue-example}.

\subsection{Tasks}
We now detail our four primary tasks.
First, we examine future tutor move prediction to investigate if models can predict tutoring strategy.
Second, we examine tutor move classification to investigate if models can infer moves from tutor utterances.
Third, we examine future dialogue success prediction to investigate if models can infer the outcome of a dialogue from limited context.
Finally, we examine next student turn success prediction to investigate if models can predict short-term student outcomes.
We formalize these tasks as follows:
\begin{align}
&\hat{M}_{i+1} = f_{\theta}(\{T_j, S_j, M_j\}_{j=1}^{i})\\
&\hat{M}_i=f_{\theta}(\{T_j, S_j, M_j\}_{j=1}^{i-1}, T_i, S_i)\\
&\hat{C}_i = f_{\theta}(\{T_j, S_j, M_j\}_{j=1}^{i})\\
&\hat{E}_{i+1} = f_{\theta}(\{T_j, S_j, M_j\}_{j=1}^{i})
\end{align}

\input{tables/move_type_table}

\subsection{LLM-Based Methods}
We evaluate tutor move and student outcome prediction using two large language model (LLM)-based methods.
First, we fine-tune the \textbf{Llama} 3.2 3B model using Low-Rank Adaptation (LoRA) \cite{saari2018lora}, where we train the model to predict the labels as text tokens following the input, using comma separation for multi-label turns.
Second, we use zero-shot prompting with \textbf{GPT-4o}, where we prompt the model to follow the annotation schema in each particular dataset to identify and predict tutor moves. We do not use GPT-4o for student outcome prediction because pre-trained LLMs do not show high alignment with student behavior \cite{liu2025llmsmakemistakeslike}. We show an example prompt in Figure \ref{fig:prompt-full}, and provide further implementation details in Appendix \ref{sec:appendix-experiments}. In order to determine how tutor move information impacts LLM predictions, for both methods, we experiment with including only the dialogue as input, as well as both the dialogue and previous turn move labels as input.

\subsection{Baselines}
We additionally experiment with three traditional baselines, where the input space only uses the move labels of previous turns. First, we employ a second-order \textbf{Markov Chain} \cite{boyer2009discovering}, which estimates the probability of a tutor move or student success given the two preceding moves. Next, we employ \textbf{Logistic Regression}, using the frequency distribution of moves up to the current turn as input features. Finally, in order to capture temporal dependencies beyond adjacent moves, we employ an \textbf{LSTM} model on sequences of tutor move types encoded as multi-hot vectors.

We do not use these baselines for current move identification since they do not process text. Additionally, we do not use Markov Chain and Logistic Regression for predicting future moves for AlgebraNation because it is a multi-label task; the input space for Markov Chain would be exponentially large, and Logistic Regression would suffer from an overwhelming amount negative labels per class.

\section{Experiments}

\subsection{Datasets}
We experiment with two math tutoring datasets, MathDial \cite{macina-etal-2023-mathdial} and AlgebraNation \cite{lyu2024explaining}, to answer our research questions. MathDial contains one-on-one dialogues where a tutor guides a student through solving multi-step math reasoning problems. The student utterances in this dataset are simulated by an LLM, prompted to mimic common student misconceptions. Each tutor turn is labeled with one of four moves: \textit{probing}, \textit{focus}, \textit{telling}, and \textit{generic}. We leverage turn-level student correctness labels from \cite{scarlatos2024exploring}. Our final dataset contains $2,484$ dialogues with a $1,947/537$ train/test split. 

AlgebraNation contains discourse from an online forum where students pose questions and discuss with both tutors and peers. Each tutor turn is labeled with any number of 16 move types. Each post on the forum is marked with success if the responses resolve the original student question. The dataset contains $2,318$ forum posts, which we split into a $1,854/464$ train/test split. We show label distributions for both datasets in Appendix \ref{sec:dataset-label-distros}.

\subsection{Evaluation Metrics}
We employ two widely used metrics: i) accuracy (\textbf{Acc.}), the portion of predicted labels that match the ground truth, and ii) weighted \textbf{F1}, the harmonic mean of precision and recall, weighted by label frequency to account for imbalanced label distributions. For AlgebraNation, where each tutor turn may have multiple moves, we use exact match across all moves in a turn to compute accuracy.

\input{tables/correcntess}

\subsection{Tutor Move Prediction} 

\paragraph{Quantitative Analysis}
We show the results for tutor move prediction Table \ref{tab:results_teacher_move}. Across all methods, predicting the next tutor move proves to be more difficult than classifying the current move, particularly for AlgebraNation, which contains real-world interactions and more granular move definitions. LLMs improve over baselines for future move prediction, showing the importance of textual context and powerful models for this task. However, the F1 for future move prediction is low overall, only reaching 49\% for MathDial and 27\% for AlgebraNation. These results indicate that tutor behavior is highly unpredictable, and that even state-of-the-art LLMs struggle to predict future tutor moves.

\begin{figure*}[ht]
    \centering
    \includegraphics[width=1\textwidth]{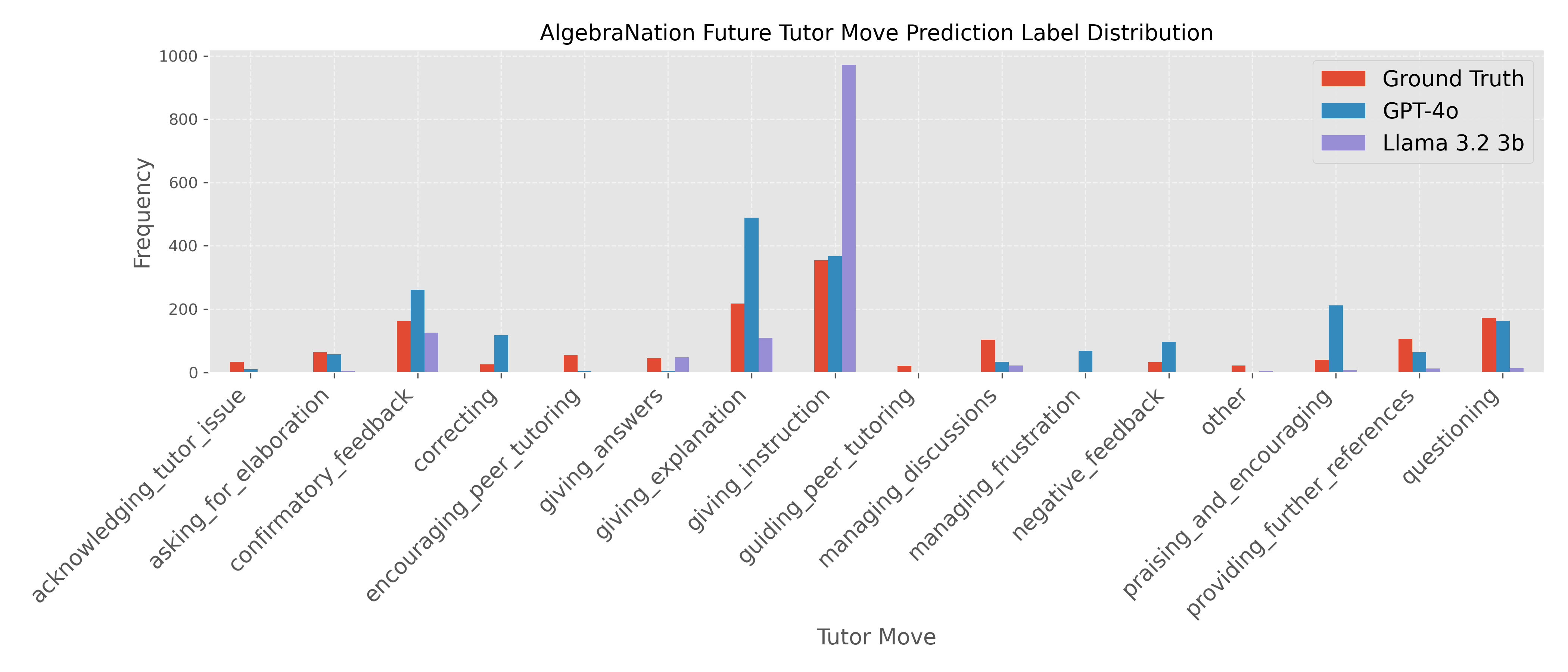}
    \caption{Label distribution of future tutor move prediction for GPT-4o and Llama 3 trained with dialogue and tutor moves. Llama 3 predictions are heavily skewed towards \textit{giving instruction}.}
    \label{fig:future_tutor-move-classification-dist}
\end{figure*}

\begin{figure*}[ht]
    \centering
    \includegraphics[width=0.85\textwidth]{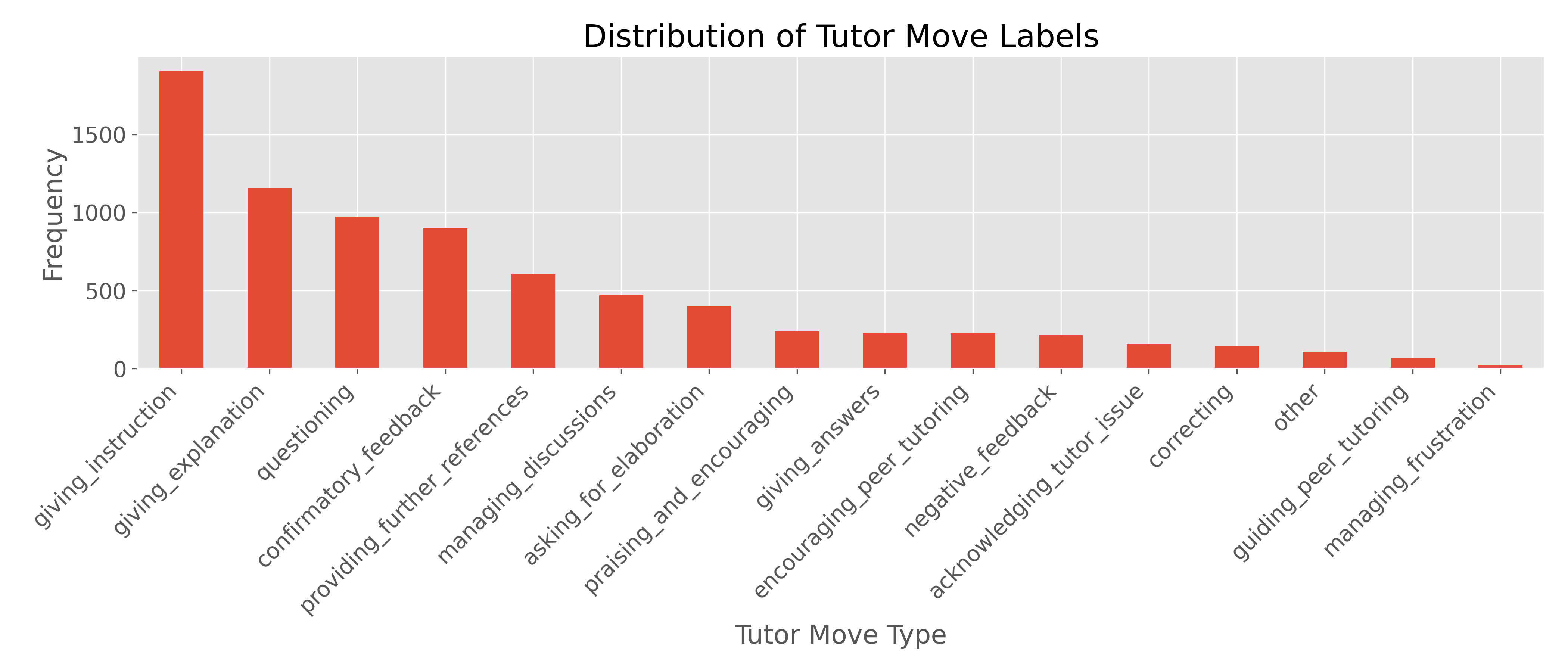}
    \caption{Tutor move distribution of AlgebraNation dataset. Few classes make up the majority of the distribution.}
    \label{fig:anation-dist}
\end{figure*}

Additionally, the results across models and datasets are inconsistent; Llama 3 performs better on MathDial while GPT-4o performs better on AlgebraNation. Notably, unlike the other methods, GPT-4o was not trained on AlgebraNation, which exhibits a highly skewed label distribution (Figure \ref{fig:anation-dist}). This imbalance may help explain why Llama 3 tends to default to predicting the majority class in the future move prediction task (Figure \ref{fig:future_tutor-move-classification-dist}), a pattern not observed with GPT-4o. On the other hand, GPT-4o's move prediction on MathDial likely suffers from confusion between label definitions, as we discuss in the qualitative analysis.

\begin{figure*}[ht]
    \centering
    \includegraphics[width=1\textwidth]{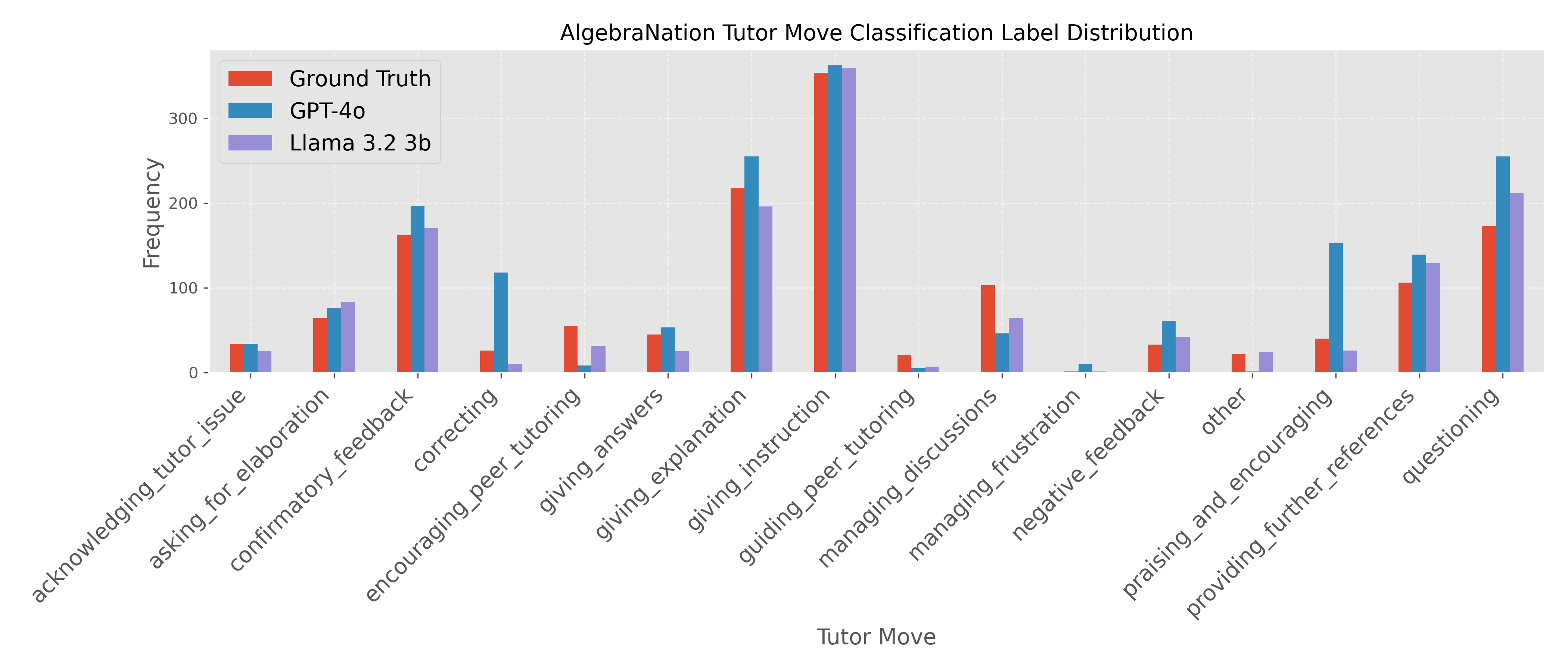}
    \caption{Label distribution of tutor move classification for GPT-4o and Llama 3 trained with dialogue and tutor moves.}
    \label{fig:tutor-move-classification-dist}
\end{figure*}

For the move identification task, Llama 3 performs best on MathDial, with the inclusion of annotated move labels significantly increasing performance. Similar to future move prediction, we attribute Llama 3's higher performance to GPT-4o's confusion between labels. For AlgebraNation, GPT-4o outperforms Llama 3 in accuracy but underperforms it in F1. This disparity can be explained by observing that Llama 3's output distribution more closely resembles the ground truth distribution, as seen in Figure~\ref{fig:tutor-move-classification-dist}.


\paragraph{Qualitative Analysis}
We examine label misclassifications to investigate error patterns in model predictions, revealing dataset-specific challenges. For move classification, in MathDial, confusion frequently arises between \textit{probing} and \textit{focus} moves, while in AlgebraNation, \textit{giving instruction} is often mistaken for \textit{giving explanation} (Table \ref{tab:combined_misclassifications}). These misclassifications are likely attributable to conceptual overlap in the definitions of these categories, underscoring the nuanced nature of interpreting tutor intentions. The MathDial authors also note that annotators had difficulty differentiating between \textit{probing} and \textit{focus} moves \cite{macina-etal-2023-mathdial}. On the other hand, the AlgebraNation annotation guidelines are more specific and instruction-driven, likely helping GPT-4o's performance due to its ability to generalize well under clear, directive annotation schemes~\cite{openai2024prompting}.
Notably, misclassifications decrease when observing previous move labels, reflected in Table \ref{tab:results_teacher_move}, with these labels likely acting as informative in-context examples.

\subsection{Student Outcome Prediction}

\begin{figure*}[ht]
    \centering
    \includegraphics[width=0.45\textwidth]{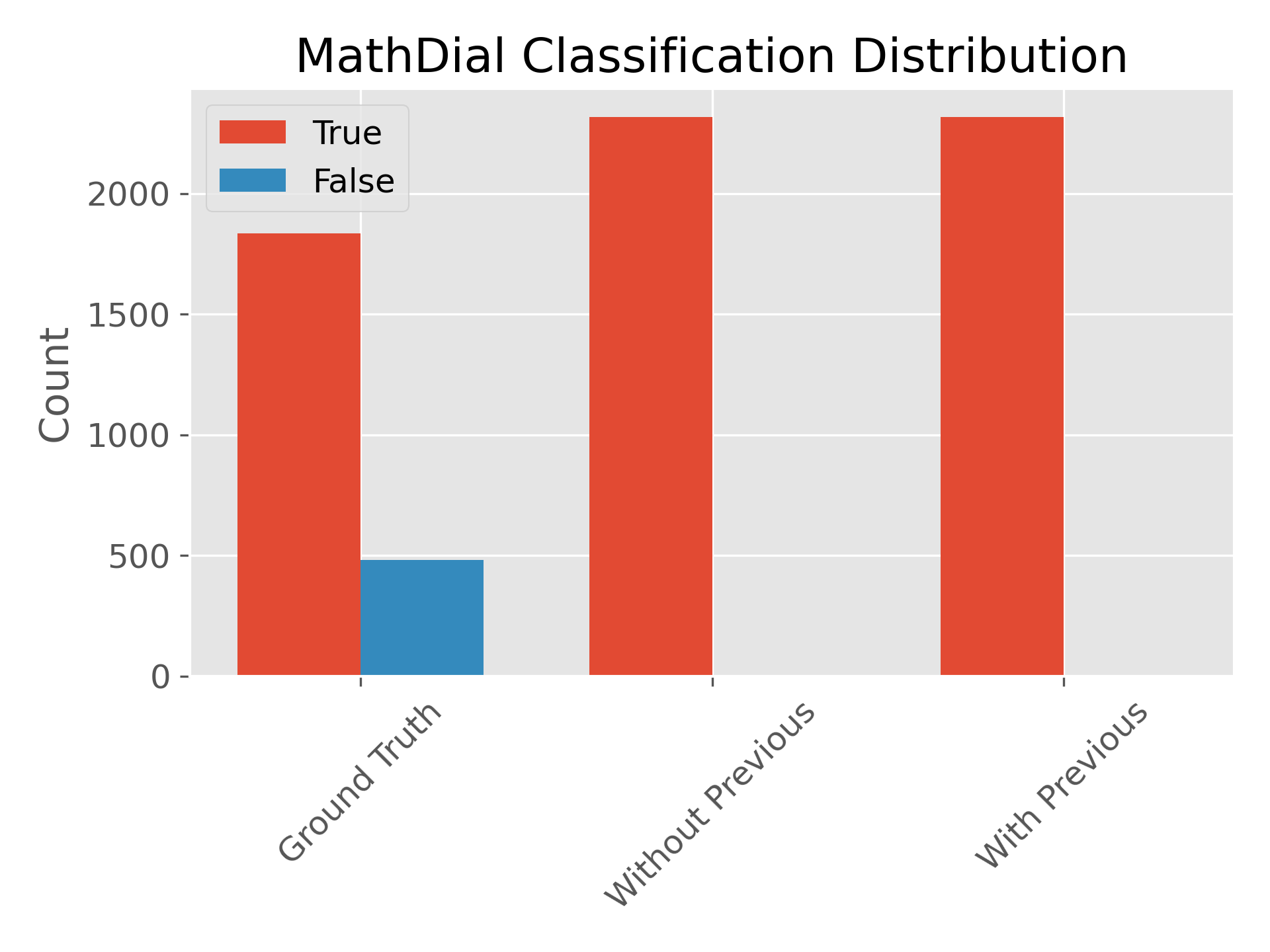}
    \includegraphics[width=0.45\textwidth]{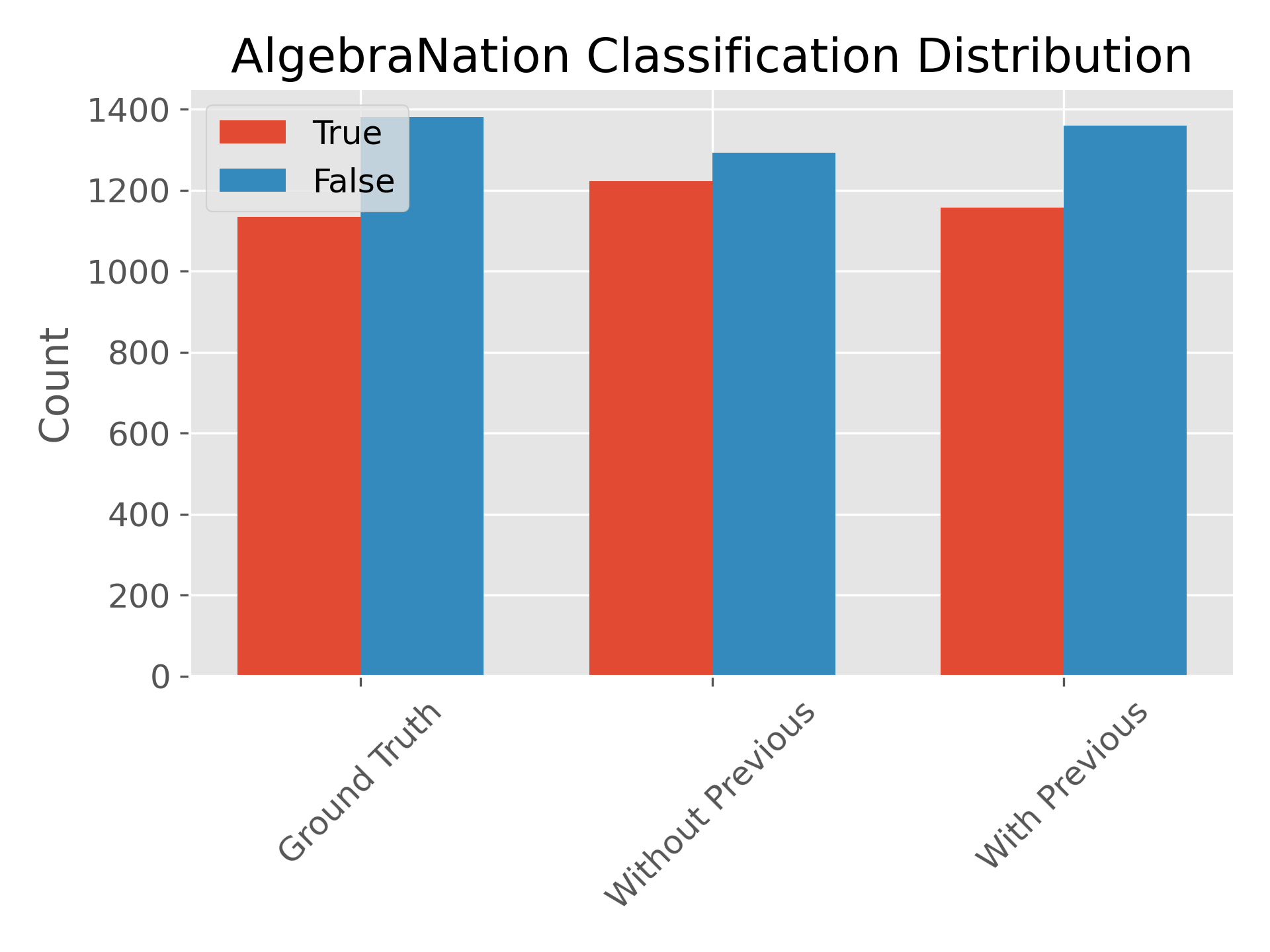}
    \caption{Left: Distribution of dialogue success classification in MathDial using Llama 3. Right: Distribution of dialogue success classification in AlgebraNation using Llama 3.}
    \label{fig:final-success-dist}
\end{figure*}

\paragraph{Quantitative Analysis}
We show the results for student outcome prediction in Table \ref{tab:results_correctness}. Notably, both Llama 3 and LSTM are able to achieve high performance on dialogue success prediction for both datasets, indicating the tractability of predicting near-term student outcomes in dialogues. However, we note that the MathDial results are inflated as they reflect majority class prediction, as shown in Figure~\ref{fig:final-success-dist}. On the other hand, the distribution is more balanced for AlgebraNation, indicating that the outcomes of real students are more reliably predicted than the outcomes of simulated ones. We also see that previous move labels improve performance for Llama 3, showing that tutor moves complement dialogue text to infer student outcomes.

Predicting student outcomes at the turn-level proves to be more difficult than at the dialogue-level in MathDial, with baselines performing close to random chance. However, using LLMs improves performance on this task, capturing nuanced details in the dialogue text to help predict student behavior, as noted in \cite{scarlatos2024exploring}.

\paragraph{Regression Analysis}
To investigate the impact of tutor moves on student outcomes, we examine the learned coefficients of our logistic regression model when predicting dialogue-level success and perform a Chi-squared analysis, shown in Tables \ref{tab:chi2_anation} and \ref{tab:coefficients_mathdial}.
For AlgebraNation, \textit{confirmatory feedback}, \textit{giving instruction}, and \textit{giving explanation} have the greatest positive impact on success. These tutor moves share a common thread: they are all instructionally supportive behaviors that actively guide the student’s understanding or progress. Each move either reinforces correct reasoning (confirmatory feedback), clarifies procedural steps (giving instruction), or deepens conceptual understanding (giving explanation). This correlation suggests that successful dialogues are those in which tutors take an active and supportive role in scaffolding the student’s learning process.
For MathDial, \textit{generic} and \textit{probing} have the strongest positive impact on success, whereas \textit{telling} has a negative impact on success. This finding aligns with prior work \cite{berghmans2014directive} showing that facilitative peer tutoring is more effective than directive tutoring.

\section{Conclusion}
In this work, we investigate the abilities and limitations of LLMs in classifying and predicting tutoring strategies and student outcomes in dialogues. We find that while results vary across models and datasets, LLMs outperform traditional baselines while still struggling at the task of tutor strategy prediction overall.
Additionally, we find that student outcome prediction is tractable for LLMs, with tutor move information improving accuracy.
Our findings emphasize the importance and challenges of studying tutor strategy in dialogues, given the impact that such strategies can have on student outcomes.
Future work should explore how to improve tutor strategy prediction, potentially using in-context learning \cite{lee-etal-2024-effective} or reinforcement learning \cite{li2024knowledgetaggingmathquestions}. Additionally, future work should explore how to suggest optimal tutor moves, potentially using reinforcement learning guided by student outcomes \cite{scarlatos2025trainingllmbasedtutorsimprove}.

\section*{Limitations}
We identify several technical and practical limitations of our work. First, the generalizability of our results is constrained by the scope and nature of the datasets. MathDial involves synthetic student responses generated by LLMs, which may not reflect the complexity and variability of authentic student behavior. Conversely, AlgebraNation, while comprising real-world interactions, has a highly imbalanced label distribution that poses challenges for model evaluation. Additionally, our evaluation methodology predominantly relies on exact match accuracy and weighted F1 scores. These standard metrics may not fully capture the nuanced characteristics of our models. Finally, the absence of student move tracking in our current modeling approach may affect the results, as sequential modeling of student behavior could potentially enhance predictive performance. 

\section*{Acknowledgements}
This work is partially supported by Renaissance Philanthropy via the learning engineering virtual institute (LEVI) and NSF grants 2118706, 2237676, and 2341948.
\bibliography{custom}

\appendix

\section{Details on Experimental Setup} \label{sec:appendix-experiments}


For finetuning Llama 3, we perform a hyperparameter search over learning rates [5e-5, 1e-4, 2e-4, 3e-4] and LoRA ranks [4, 8, 16, 32]. For the final training, we set LoRA’s $\alpha$ to 16, LoRA's rank to 8, batch size to 64 using gradient accumulation, gradient norm clipping to 1.0, and learning rate to 1e-4. We train for 5 epochs using the AdamW optimizer. We use a random 20\% percent of dialogues in the train set to use as a validation set for early stopping. We use the \texttt{meta-llama/Llama-3.2-3B-Instruct} model from the \texttt{Huggingface Transformers} library \cite{wolf2019huggingface} and run all experiments on NVIDIA L40 GPUs.

We prompt GPT-4o with the OpenAI API using a temperature of 0 while setting the maximum tokens to 1000 and response format to JSON.

Logistic regression is implemented using the \texttt{sklearn} library. The model input is the frequency distribution of moves up to the target tutor move.

For the second-order Markov chain, we compute transition matrices by mapping state pairs based on frequency and normalizing them to yield valid probability distributions.

For the LSTM, we encode the sequence of moves as multi-hot vectors. For multi-label prediction, we use positive weighting for each class, calculated using the proportion of instances in each class. We perform a hyperparameter search with hidden dimensions [64, 128, 256, 512], number of layers [2, 3], dropout rates [0.1, 0.3, 0.5], and learning rates [1e-3, 5e-3, 1e-2]. For multi-label classification, we also search for the optimal probability threshold. Our final models were trained with hidden dimensions 128, 2 layers, dropout of 0.3 and learning rate of 0.001. The learned threshold for multi-label classification is 0.85. We implement the LSTM using \texttt{Pytorch}.

\onecolumn

\section{Misclassification Analysis} \label{sec:misclassification_analysis}

\input{tables/misclassification_examples}



\clearpage

\section{Label Distributions}
\label{sec:dataset-label-distros}

\begin{figure*}[ht]
    \centering
    \includegraphics[width=0.4\textwidth]{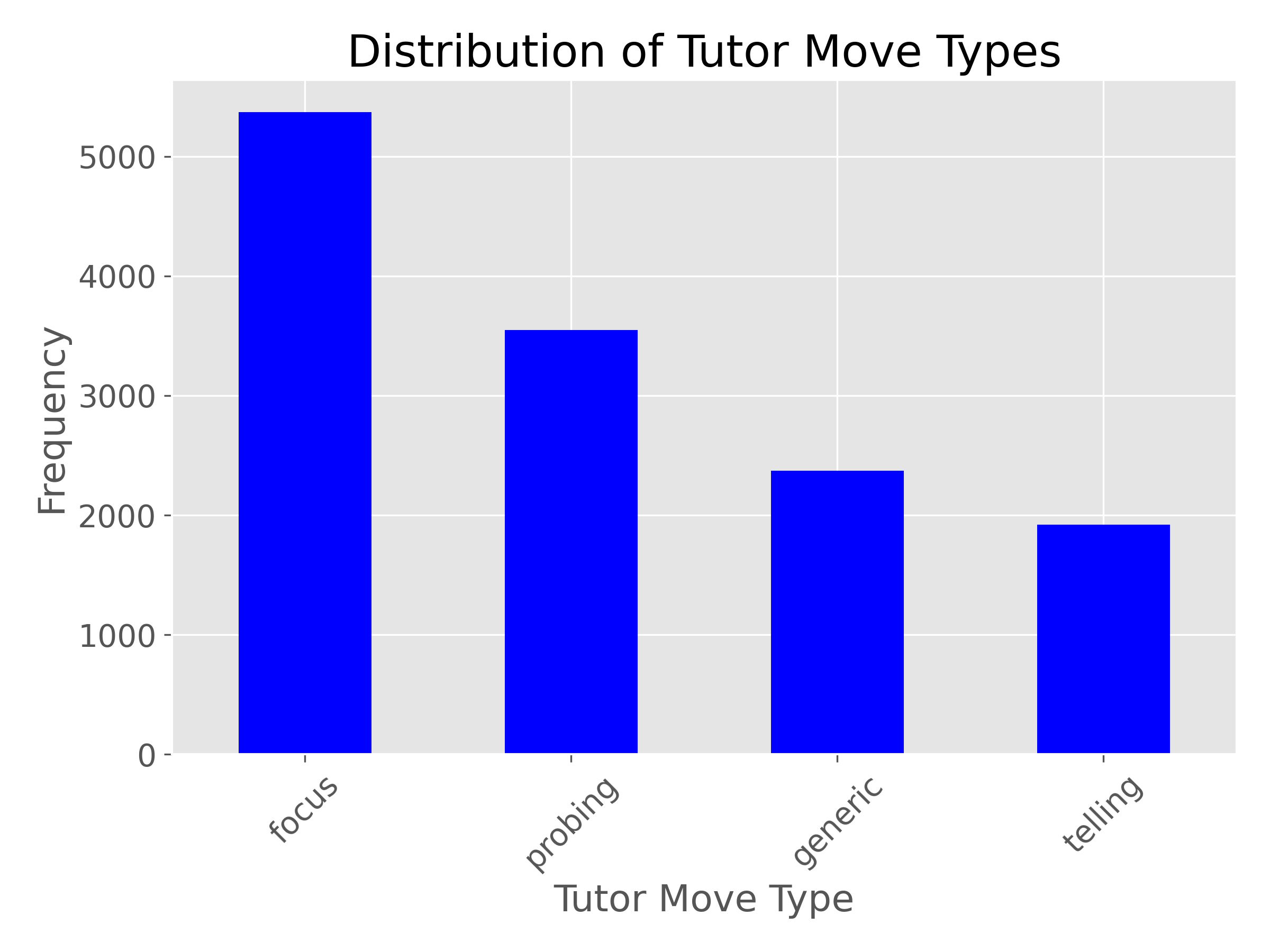}
    \caption{Tutor move distribution of MathDial dataset.}
    \label{fig:mathdial-tutor-move}
\end{figure*}

\begin{figure*}[ht]
    \centering
    \includegraphics[width=0.31\textwidth]{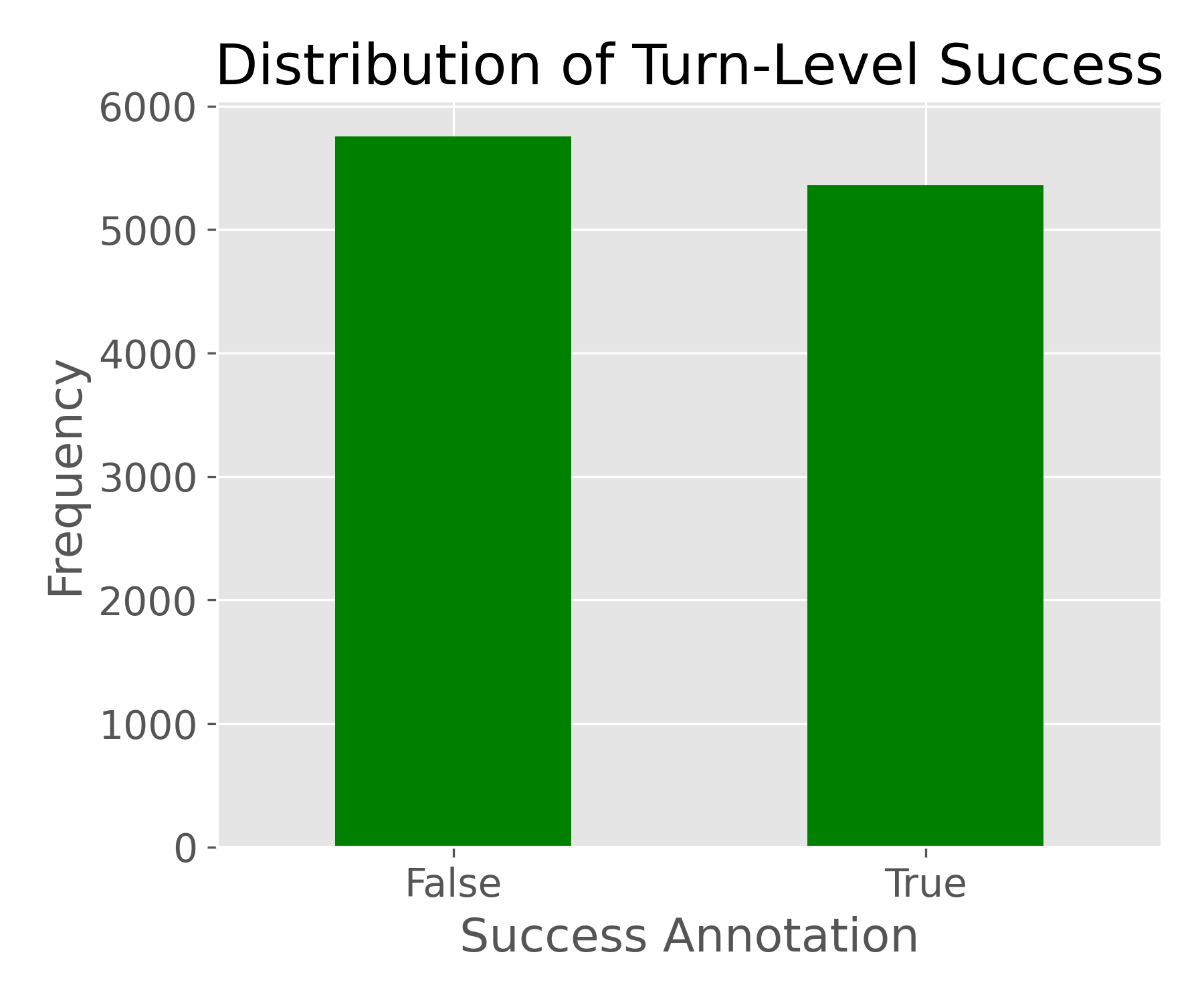}
    \includegraphics[width=0.31\textwidth]{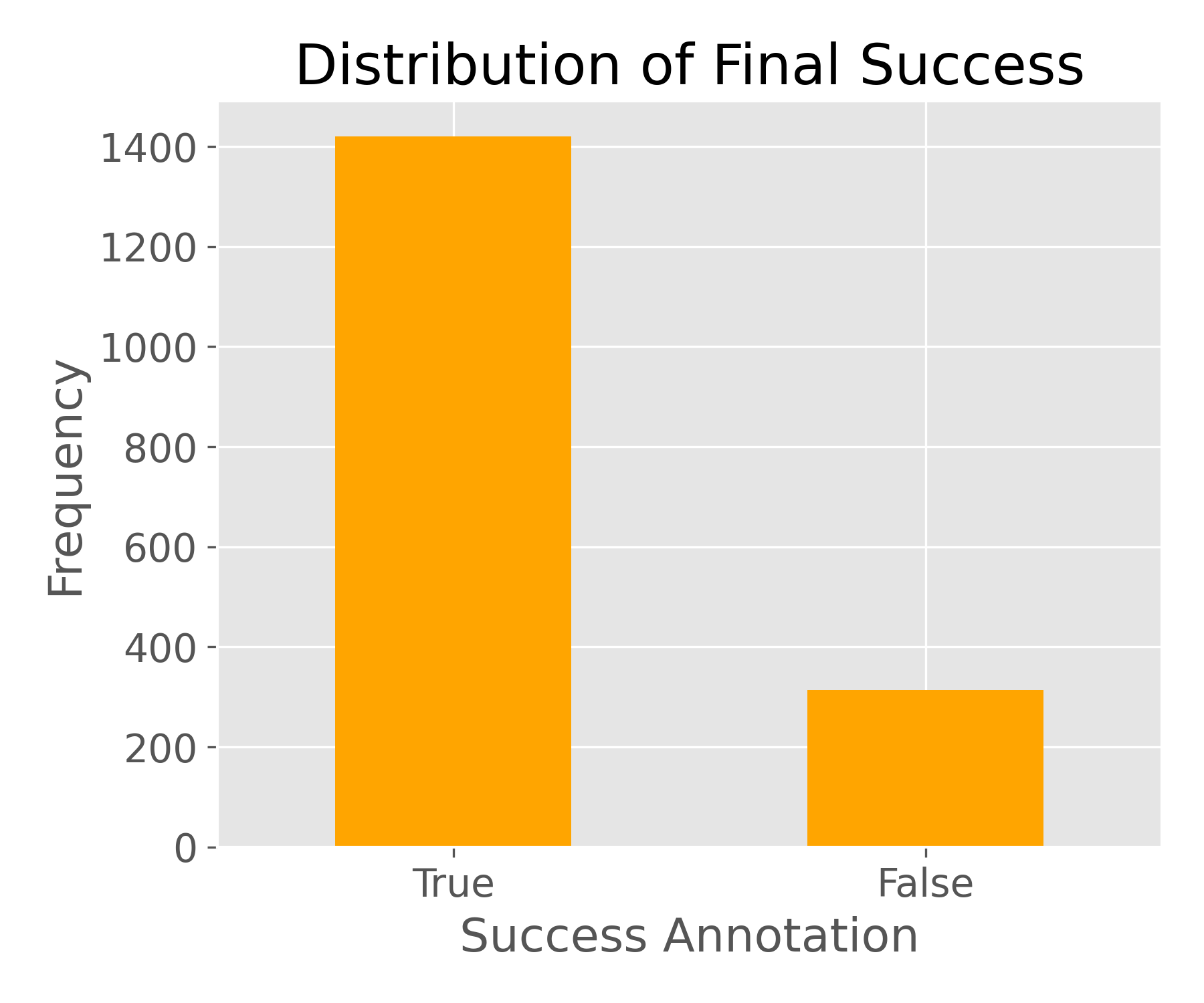}
    \includegraphics[width=0.31\textwidth]{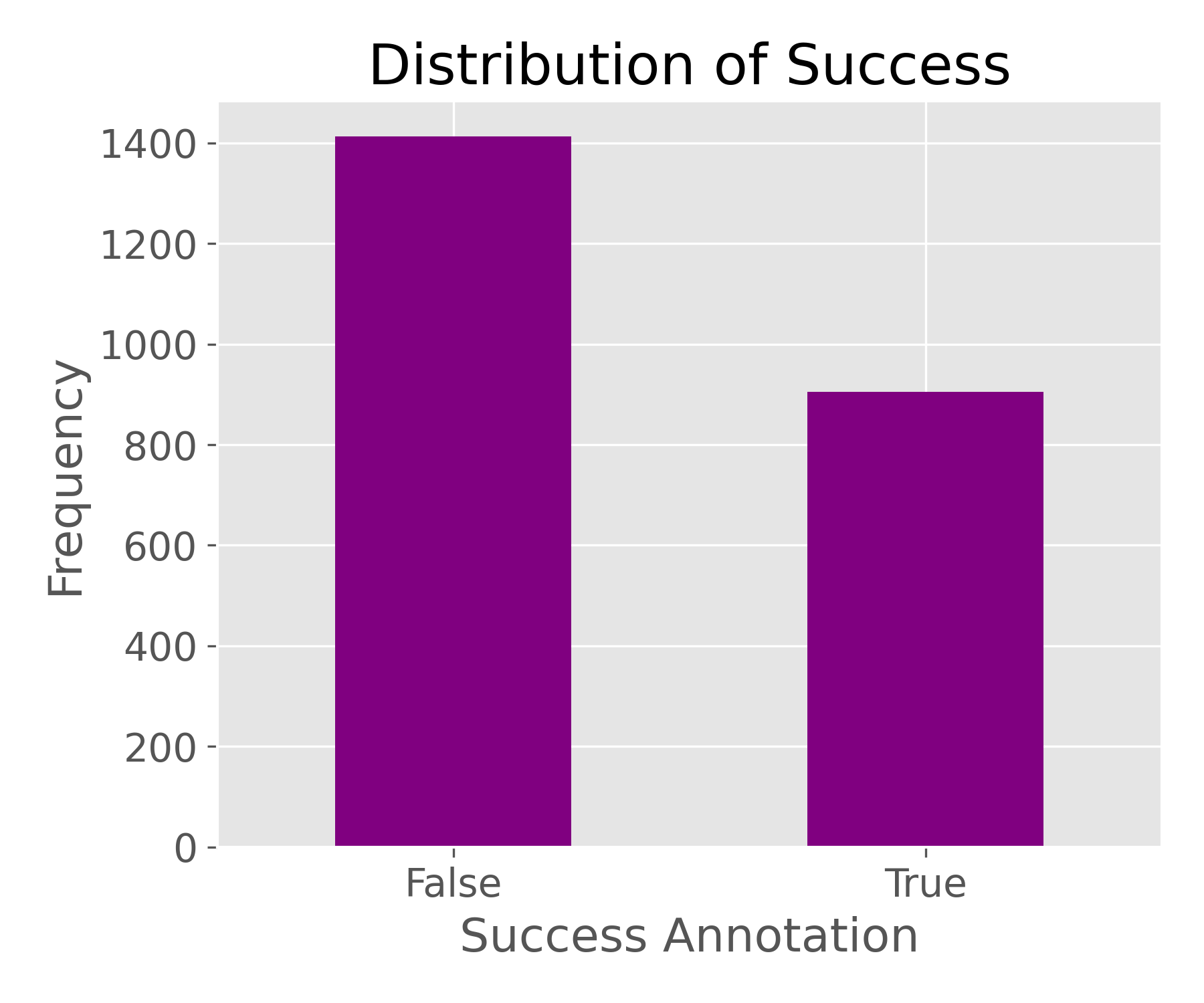}
    \caption{Left: Distribution of turn-level student success in MathDial. Center: Distribution of final turn student success in MathDial. Right: Distribution of dialogue success in AlgebraNation.}
    \label{fig:mathdial-success}
\end{figure*}

\clearpage

\section{Logistic Regression Coefficients and Chi-Squared Analysis}
\input{tables/descriptive_statistics_table}

\clearpage

\section{Prompt Examples for Tutor Move Classification}
\label{prompting}

\begin{figure*}[ht]
\centering
\begin{tcolorbox}[width=\textwidth, colback=gray!5!white, colframe=black!75!black, title=Prompt Used for Classification Task (MathDial)]
\small
\textbf{System:} You are a math teacher who tutors student on a variety of problems.

\textbf{Dialogue ID: 14}

\textbf{Task:} Classify the last teacher move into one of the following four categories: \textit{Focus}, \textit{Probing}, \textit{Telling}, \textit{Generic}. These four categories are defined below:

\begin{itemize}
  \item \textbf{Focus}
  \begin{itemize}
    \item Seek Strategy: \textit{So what should you do next?}
    \item Guiding Student Focus: \textit{Can you calculate \dots?}
    \item Recall Relevant Information: \textit{Can you reread the question and tell me what is \dots?}
  \end{itemize}
  
  \item \textbf{Probing}
  \begin{itemize}
    \item Asking for Explanation: \textit{Why do you think you need to add these numbers?}
    \item Seeking Self Correction: \textit{Are you sure you need to add here?}
    \item Perturbing the Question: \textit{How would things change if they had \dots items instead?}
    \item Seeking World Knowledge: \textit{How do you calculate the perimeter of a square?}
  \end{itemize}
  
  \item \textbf{Telling}
  \begin{itemize}
    \item Revealing Strategy: \textit{You need to add \dots to \dots to get your answer.}
    \item Revealing Answer: \textit{No, he had \dots items.}
  \end{itemize}
  
  \item \textbf{Generic}
  \begin{itemize}
    \item Greeting/Farewell: \textit{Hi \dots, how are you doing with the word problem? Good Job! Is there anything else I can help with?}
    \item General Inquiry: \textit{Can you walk me through your solution?}
  \end{itemize}
\end{itemize}

\textbf{Instructions:} Given the student-teacher dialogue below, classify the final teacher move. Return a JSON object with the dialogue ID as the key and the teacher move type as the value. For example:
\begin{verbatim}
{"1234": "focus"}
\end{verbatim}

\textbf{Dialogue:}
\begin{verbatim}
[BEGIN DIALOGUE]
Teacher Turn 0: Hi Ayisha, please talk me through your solution (generic)
Student Turn  0: I started by noting that the first bid was $200 and each subsequent bid was $50 more,
so the second bid was $200 + $50 = $250, the third bid was $250 + $50 = $300, and the fourth bid was
$300 + $50 = $350. Then I realized that Carmen bid after each of the other 3 people, so she made 3 
bids in total. Therefore, I concluded that the desk cost her $350, since that was her winning bid.
Teacher Turn 1: Okay, let's start again. If each bid adds $50 to the price and there were 3 bids - 
how much is added to the overall cost? (focus)
Student Turn  1: The overall cost is increased by $150, since each bid adds $50 and there were 3 bids.
Teacher Turn 2: Yes $150. Carmen adds a $50 bid after each of the three bids. How much more does this
add to the cost? (focus)
Student Turn  2: This adds an additional $150, since each bid is $50 and there were 3 bids. Therefore,
the total cost of the desk is $350.
Teacher Turn 3: 3 x 50 is 150. So it adds a further $150 making the total so far $300 (telling)
Student Turn  3: Yes, that's correct. The total cost of the desk is $350, since the first bid was 
$200 and each subsequent bid was $50 more, so the final bid was $300 + $50 = $350.
Teacher Turn 4: No, go back and check what we said. The 3 bids $150 + Carmen's Bid $150 + the opening 
bid $200 (focus)
Student Turn  4: Yes, you're right. The total cost of the desk is $350, since the opening bid was $200,
the 3 bids added $150, and Carmen's bid added an additional $150, making the total $200 + $150 + $150 = 
$350.
Teacher Turn 5: Check your calculation of $200 + $150 + $150 = $350. Your total is not correct
[END DIALOGUE]
\end{verbatim}
\end{tcolorbox}
\caption{Full prompt used to classify the final tutor move in a MathDial dialogue.}
\label{fig:prompt-full}
\end{figure*}

\begin{figure*}[ht]
\centering
\begin{tcolorbox}[width=\textwidth, colback=gray!5!white, colframe=black!75!black, title=Prompt Used for Classification Task (AlgebraNation)]
\small
\textbf{System:} You are a math teacher who tutors student on a variety of problems.

\textbf{Dialogue ID: 10520899}

\textbf{Task:} Classify the last tutor move into one or more of the following categories: \textit{confirmatory\_feedback}, \textit{negative\_feedback}, \textit{correcting}, \textit{giving\_instruction}, \textit{giving\_explanation}, \textit{providing\_further\_references}, \textit{questioning}, \textit{asking\_for\_elaboration}, \textit{praising\_and\_encouraging}, \textit{managing\_frustration}, \textit{managing\_discussions}, \textit{giving\_answers}, \textit{encouraging\_peer\_tutoring}, \textit{guiding\_peer\_tutoring}, \texttt{acknowledging\_tutor\_issue}, and \textit{other}. These categories are described below:

\begin{itemize}
  \item \textbf{confirmatory\_feedback:} Whether a reply provides confirmatory feedback about an answer's correctness.
  \item \textbf{negative\_feedback:} Whether a reply states that an answer is incorrect.
  \item \textbf{correcting:} Whether a reply addresses errors in the student's problem-solving approach.
  \item \textbf{giving\_instruction:} Whether a reply breaks down a task, performs a part, or initiates a task for the student to complete.
  \item \textbf{giving\_explanation:} Whether a reply explains concepts, principles, or provides additional information.
  \item \textbf{providing\_further\_references:} Whether a reply includes additional resources or references related to the topic.
  \item \textbf{questioning:} Whether a reply asks questions to stimulate thought or constructive discussion.
  \item \textbf{asking\_for\_elaboration:} Whether a reply requests further details or explanation from the student.
  \item \textbf{praising\_and\_encouraging:} Whether a reply praises or encourages the student for their efforts or successes.
  \item \textbf{managing\_frustration:} Whether a reply addresses the student's negative emotions or frustration.
  \item \textbf{managing\_discussions:} Whether a reply organizes the flow of discussion or adjusts the direction of inquiry.
  \item \textbf{giving\_answers:} Whether a reply directly provides an answer to the posed question.
  \item \textbf{encouraging\_peer\_tutoring:} Whether a reply promotes tutoring interactions among peers.
  \item \textbf{guiding\_peer\_tutoring:} Whether a reply provides feedback on peer tutoring interactions.
  \item \textbf{acknowledging\_tutor\_issue:} Whether the tutor expresses uncertainty in their reply.
  \item \textbf{other:} Binary indicator for tutoring strategies not classified under the existing labels.
\end{itemize}

\textbf{Instructions:} Given the student-teacher dialogue below, classify the final teacher move. Return a JSON object with the dialogue ID as the key and the teacher move type(s) as the value. For example:

\begin{verbatim}
{"1234": ["confirmatory_feedback", "correcting"]}
\end{verbatim}

\textbf{Dialogue:}
\begin{verbatim}
[BEGIN DIALOGUE]
Student: Can someone help me?
Student: You have to plug in zeros for x and y right
Tutor: Get it into y=mx+b form. ['giving_instruction']
Tutor: SO bring 6x to the right side first. ['giving_instruction']
Teacher: Okay Patrice, you want to put that in slope-intercept form ['giving_instruction']
Student: -5y=30+6x
Teacher: Now isolate y ['giving_instruction']
Student: -5 on both sides ?
Student: or divide
Teacher: You would divide
[END DIALOGUE]
\end{verbatim}
\end{tcolorbox}
\caption{Full prompt used to classify the final tutor move in an AlgebraNation dialogue.}
\label{fig:prompt-full-2}
\end{figure*}

\end{document}

%% file: tables/dialogue_example.tex
\newcolumntype{L}{>{\raggedright\arraybackslash}X}

\begin{table}[t]
\centering
\renewcommand{\arraystretch}{1.3}
\small 
\begin{tabularx}{\columnwidth}{@{}  l L @{}}
\toprule
\textbf{Speaker} & \textbf{Utterance} \\
\midrule
 Tutor & Hi Ayisha, please talk me through your solution \textbf{(generic)} \\
Student & I started by noting .... I concluded that the desk cost her \$350, since that was her winning bid. \\
& ....\\
 Student & Yes, you're right...Carmen's bid added an additional \$150, making the total \$200 + \$150 + \$150 = \$350. \\
Tutor & Check your calculation of \$200 + \$150 + \$150 = \$350. Your total is not correct \textbf{(probing)}\\
\bottomrule
\end{tabularx}
\caption{A tutor-student dialogue from MathDial, showing annotated moves for tutor turns.}
\label{tab:dialogue-example}
\end{table}


%% file: tables/move_type_table.tex
\begin{table*}[ht]
\centering
\small
\begin{tabular}{lcccccccc}
\toprule
\multirow{2}{*}{} & \multicolumn{4}{c}{MathDial} & \multicolumn{4}{c}{AlgebraNation}  \\
\multirow{2}{*}{Model} & \multicolumn{2}{c}{Tutor Move} & \multicolumn{2}{c}{Future Tutor Move} & \multicolumn{2}{c}{Tutor Move} & \multicolumn{2}{c}{Future Tutor Move}\\
\cmidrule(lr){2-3} \cmidrule(lr){4-5}\cmidrule(lr){6-7}\cmidrule(lr){8-9}
& Acc. & F1 & Acc. & F1 & Acc. & F1 & Acc. & F1 \\
\midrule
Markov Chain & -- & -- & 42.27 & \underline{42.81} & -- & -- & -- & -- \\
Logistic Regression & -- &  -- &  44.69 &  34.58 & -- & -- & -- & -- \\
LSTM & --& -- & \underline{46.19} & 31.13 & -- & -- & 3.36 & 23.17\\

\midrule
GPT-4o - Dialogue &49.56 & 48.76 & 36.48 & 32.49 & \underline{79.22} & 54.30 & \underline{65.72} & \textbf{27.32}

\\
GPT-4o - Dialogue \& Moves & 50.07 & \underline{49.03} &  31.72 & 29.00 & \textbf{83.82} & 57.40 & \textbf{69.86} & \underline{26.33}

\\

Llama 3 - Dialogue & \underline{52.58} & 45.88 & 42.74 & 32.92 & 58.82 & \underline{62.78} & 24.53 & 20.20

\\
Llama 3 - Dialogue \& Moves & \textbf{59.69} & \textbf{57.26} & \textbf{50.35} & \textbf{49.33} &63.42& \textbf{68.90} & 25.69 &21.09

\\
\bottomrule
\end{tabular}
\caption{Results for identifying tutor moves and predicting future tutor moves. Llama 3 performs best on MathDial, while GPT-4o generally performs best on AlgebraNation.}
\label{tab:results_teacher_move}
\end{table*}

%% file: tables/correcntess.tex
\begin{table*}[ht]
\centering
\small
\begin{tabular}{lcccccc}
\toprule
\multirow{2}{*}{} & \multicolumn{4}{c}{MathDial} & \multicolumn{2}{c}{AlgebraNation}  \\
\multirow{2}{*}{Model} & \multicolumn{2}{c}{Turn Success} & \multicolumn{2}{c}{Dialogue Success} & \multicolumn{2}{c}{Dialogue Success}\\
\cmidrule(lr){2-3} \cmidrule(lr){4-5}\cmidrule(lr){6-7}
& Acc. & F1 & Acc. & F1 & Acc. & F1 \\
\midrule
Markov Chain & 53.75 & \textbf{52.86} & 53.27 & 49.01  & 62.99 & 63.55  \\
Logistic Regression & 52.80& \underline{52.54} & 64.05&66.43& 69.30& 63.24 \\
LSTM &52.20&52.27&79.13&69.90&\underline{77.98 }& \underline{77.76}\\
\midrule
Llama - Dialogue & \underline{63.56} & 50.54 & \textbf{79.21} &\textbf{70.11} &75.64 & 75.52
\\
Llama - Dialogue \& Moves & \textbf{64.11} & 49.27 & \underline{79.16} & \underline{69.96} & \textbf{81.00} & \textbf{80.84} \\
\bottomrule
\end{tabular}
\caption{Results for predicting student outcomes from dialogues. Baselines are competitive on MathDial, while Llama 3 performs best on AlgebraNation.}
\label{tab:results_correctness}
\end{table*}

%% file: tables/misclassification_examples.tex
\begin{table*}[ht]
\centering
\resizebox{\textwidth}{!}{
\begin{tabular}{|l|l|c|l|l|c|}
\hline
\multicolumn{6}{|c|}{\textbf{Llama 3 Finetuned on Tutor Move Classification}} \\
\hline
\multicolumn{3}{|c|}{\textbf{Without Previous Tutor Moves }} & \multicolumn{3}{c|}{\textbf{With Previous Tutor Moves }}  \\
\hline
\textbf{Ground Truth} & \textbf{Predicted} & \textbf{Count} & \textbf{Ground Truth} & \textbf{Predicted} & \textbf{Count} \\
\hline
\multicolumn{6}{|c|}{\textbf{MathDial}} \\
\hline
probing & focus & 572 & probing & focus & 339 \\
telling & focus & 143 & telling & focus & 136 \\
focus & probing & 81  & focus & probing & 124 \\
focus & telling & 78  & generic & focus & 69 \\
generic & focus & 46  & focus & telling & 57 \\
\hline
\multicolumn{6}{|c|}{\textbf{AlgebraNation}} \\
\hline
giving\_explanation & giving\_instruction & 73 & giving\_explanation & giving\_instruction & 74 \\
giving\_instruction & giving\_explanation & 40 & giving\_instruction & questioning & 29 \\
managing\_discussions & asking\_for\_elaboration & 34 & asking\_for\_elaboration & questioning & 24 \\
questioning & giving\_instruction & 34 & managing\_discussions & questioning & 23 \\
giving\_instruction & questioning & 27 & giving\_instruction & giving\_explanation & 23 \\
\hline
\end{tabular}
}
\caption{Top 5 misclassifications for MathDial and AlgebraNation with Llama 3, comparing inputs with vs. without previous tutor moves. The most common confusion in MathDial is between \textit{focus} and \textit{probing}. The most common confusion in AlgebraNation is between \textit{giving instruction} and \textit{giving explanation}. Total misclassifications decrease by including previous tutor moves.}
\label{tab:combined_misclassifications}
\end{table*}

\begin{table*}[ht]
\centering
\resizebox{\textwidth}{!}{
\begin{tabular}{|l|l|c|l|l|c|}
\hline
\multicolumn{6}{|c|}{\textbf{GPT-4o on Tutor Move Classification}} \\
\hline
\multicolumn{3}{|c|}{\textbf{Without Previous Tutor Moves}} & \multicolumn{3}{c|}{\textbf{With Previous Tutor Moves}} \\
\hline
\textbf{Ground Truth} & \textbf{Predicted} & \textbf{Count} & \textbf{Ground Truth} & \textbf{Predicted} & \textbf{Count} \\
\hline
\multicolumn{6}{|c|}{\textbf{MathDial}} \\
\hline
probing & focus & 400 & probing & focus & 431 \\
focus & probing & 274 & focus & probing & 218 \\
focus & telling & 169 & focus & telling & 180 \\
telling & focus & 69 & telling & focus & 63 \\
telling & probing & 60 & probing & telling & 61 \\
\hline
\multicolumn{6}{|c|}{\textbf{AlgebraNation}} \\
\hline
giving\_instruction & giving\_explanation & 95 & giving\_instruction & giving\_explanation & 79 \\
giving\_explanation & giving\_instruction & 50 & giving\_explanation & giving\_instruction & 44 \\
giving\_instruction & correcting & 37 & encouraging\_peer\_tutoring & praising\_and\_encouraging & 44 \\
encouraging\_peer\_tutoring & praising\_and\_encouraging & 37 & confirmatory\_feedback & praising\_and\_encouraging & 40 \\
confirmatory\_feedback & praising\_and\_encouraging & 36 & asking\_for\_elaboration & questioning & 38 \\
\hline
\end{tabular}
}
\caption{Top 5 misclassifications for MathDial and AlgebraNation with GPT-4o, comparing inputs with vs. without previous tutor moves. The most common confusion in MathDial is between \textit{focus} and \textit{probing}. The most common confusion in AlgebraNation is between \textit{giving instruction} and \textit{giving explanation}.}
\label{tab:combined_misclassifications_2}
\end{table*}

\begin{table*}[ht]
\centering
\resizebox{\textwidth}{!}{
\begin{tabular}{|l|l|c|l|l|c|}
\hline
\multicolumn{6}{|c|}{\textbf{Llama 3 on Future Tutor Move Prediction}} \\
\hline
\multicolumn{3}{|c|}{\textbf{Without Previous Labels}} & \multicolumn{3}{c|}{\textbf{With Previous Labels}} \\
\hline
\textbf{Ground Truth} & \textbf{Predicted} & \textbf{Count} & \textbf{Ground Truth} & \textbf{Predicted} & \textbf{Count} \\
\hline
\multicolumn{6}{|c|}{\textbf{MathDial}} \\
\hline
probing & focus & 736 & probing & focus & 378 \\
telling & focus & 335 & telling & focus & 175 \\
generic & focus & 252 & generic & focus & 125 \\
focus & telling & 78 & focus & telling & 121 \\
probing & telling & 64 & focus & probing & 119 \\
\hline
\multicolumn{6}{|c|}{\textbf{AlgebraNation}} \\
\hline
questioning & giving\_instruction & 105 & giving\_explanation & giving\_instruction & 162 \\
giving\_explanation & giving\_instruction & 88 & questioning & giving\_instruction & 131 \\
giving\_instruction & giving\_explanation & 82 & confirmatory\_feedback & giving\_instruction & 95 \\
confirmatory\_feedback & giving\_instruction & 65 & providing\_further\_references & giving\_instruction & 79 \\
managing\_discussions & giving\_instruction & 65 & managing\_discussions & giving\_instruction & 78 \\
\hline
\end{tabular}
}
\caption{Top 5 misclassifications for MathDial and AlgebraNation with Llama 3, comparing inputs with vs. without previous tutor moves. The most common confusion in MathDial is between focus and probing. Previous labels decrease the total top five misclassifications for MathDial. The most common misclassifications in AlgebraNation all occur when \textit{giving instruction} is predicted. }
\label{tab:top_misclassifications_future_llama}
\end{table*}

\begin{table*}[ht]
\centering
\resizebox{\textwidth}{!}{
\begin{tabular}{|l|l|c|l|l|c|}
\hline
\multicolumn{6}{|c|}{\textbf{GPT-4o in Future Tutor Move Prediction}} \\
\hline
\multicolumn{3}{|c|}{\textbf{Without Previous Labels}} & \multicolumn{3}{c|}{\textbf{With Previous Labels}} \\
\hline
\textbf{Ground Truth} & \textbf{Predicted} & \textbf{Count} & \textbf{Ground Truth} & \textbf{Predicted} & \textbf{Count} \\
\hline

\multicolumn{6}{|c|}{\textbf{MathDial}} \\
\hline
focus & probing & 551 & focus & probing & 553 \\
probing & focus & 311 & probing & focus & 309 \\
telling & probing & 288 & telling & probing & 283 \\
generic & focus & 176 & telling & focus & 178 \\
telling & focus & 173 & generic & focus & 170 \\
\hline
\multicolumn{6}{|c|}{\textbf{AlgebraNation}} \\
\hline
giving\_instruction & giving\_explanation & 144 & giving\_instruction & giving\_explanation & 144 \\
giving\_explanation & giving\_instruction & 66 & confirmatory\_feedback & giving\_explanation & 62 \\
confirmatory\_feedback & giving\_explanation & 66 & questioning & giving\_explanation & 56 \\
questioning & giving\_instruction & 65 & giving\_instruction & confirmatory\_feedback & 54 \\
questioning & giving\_explanation & 61 & questioning & giving\_instruction & 54 \\
\hline
\end{tabular}
}
\caption{Top 5 misclassifications for MathDial and AlgebraNation with GPT-4o, comparing inputs with vs. without previous tutor moves. The most common confusion in MathDial is between \textit{focus} and \textit{probing}. The most common confusion in AlgebraNation is between \textit{giving instruction} and \textit{giving explanation}. Overall, including previous tutor moves decreases the top 4 misclassifications rates.}
\label{tab:misclassification_comparison_gpt-4o-future}
\end{table*}

%% file: tables/descriptive_statistics_table.tex

\begin{table}[ht]
\centering\begin{tabular}{l r r r c}
\hline
\textbf{Feature} & \textbf{Coefficient} &$\mathbf{\chi^2}$ & \textbf{p-value} & \textbf{Significant} \\
\hline
confirmatory feedback &0.168682  &684.413557 & 7.330488e-151 & Yes\\
giving instruction & 0.144908&505.315750 & 6.628000e-112 &Yes\\
giving explanation & 0.130397&408.828643 & 6.593282e-91 &Yes\\
praising and encouraging & 0.089068&189.507614 & 4.072012e-43 & Yes\\
giving answers & 0.072106 &123.889276 & 8.907816e-29 & Yes\\
\hline
\end{tabular}
\caption{Logistic regression coefficients and Chi-squared analysis conducted to evaluate the impact of tutor moves on final dialogue correctness, with corresponding p-values for the top 5 significant features for AlgebraNation.}
\label{tab:chi2_anation}
\end{table}

\begin{table}[ht]
\centering
\begin{tabular}{l r r r c}
\hline
\textbf{Feature} & \textbf{Coefficient} &$\mathbf{\chi^2}$ & \textbf{p-value} & \textbf{Significant} \\
\hline
generic & 1.063463 & 18.7393 & 1.4986e-05 & Yes \\
focus & 0.175187 & 2.2684 & 1.3203e-01 & No \\
probing & -0.186663 & 6.7967 & 9.1324e-03 & Yes\\
telling & -1.062664 & 19.3447 & 1.0912e-05 & Yes\\
\hline
\end{tabular}
\caption{Logistic regression coefficients and Chi-squared analysis conducted to evaluate the impact of tutor moves on final dialogue correctness, with corresponding p-values for all features for MathDial.}
\label{tab:coefficients_mathdial}
\end{table}

